\documentclass[letterpaper, 10 pt, conference]{ieeeconf}  % Comment this line out if you need a4paper

\IEEEoverridecommandlockouts                              % This command is only needed if 
\usepackage{color}
\usepackage{amsmath}
\usepackage{amsfonts}
\usepackage{bbm}
\usepackage[dvipsnames]{xcolor}
\usepackage{subfig}
\usepackage{makecell}
\usepackage{graphicx}
\usepackage{multirow}
\usepackage{wrapfig}
\usepackage{cite}
\usepackage{makecell}
\usepackage{comment}
\usepackage{cleveref}

\newcommand{\bb}[1]{\mathbb{#1}}
\newcommand{\bs}[1]{\boldsymbol{#1}} % Expectation
\title{\LARGE \bf
Symphony: Learning Realistic and Diverse\\Agents for Autonomous Driving Simulation}

\author{Maximilian Igl$^{1}$, Daewoo Kim$^{1}$, Alex Kuefler$^{1}$, Paul Mougin$^{1}$, Punit Shah$^{1}$, Kyriacos Shiarlis$^{1}$,\\Dragomir Anguelov$^{2}$, Mark Palatucci$^{2}$, Brandyn White$^{2}$, Shimon Whiteson$^{2}$

\thanks{$^{1}$Contributing author (listed alphabetically).
        $^{2}$Senior author (listed alphabetically). All authors from Waymo Research.
        Contact: {\tt\small shimonw@waymo.com}}%
}

\begin{document}

\maketitle
\thispagestyle{empty}
\pagestyle{empty}

%%%%%%%%%%%%%%%%%%%%%%%%%%%%%%%%%%%%%%%%%%%%%%%%%%%%%%%%%%%%%%%%%%%%%%%%%%%%%%%%
\begin{abstract}
Simulation is a crucial tool for accelerating the development of autonomous vehicles.  Making simulation realistic requires models of the human road users who interact with such cars.  Such models can be obtained by applying \emph{learning from demonstration} (LfD) to trajectories observed by cars already on the road.  However, existing LfD methods are typically insufficient, yielding policies that frequently collide or drive off the road. To address this problem, we propose \emph{Symphony}, which greatly improves realism by combining conventional policies with a \emph{parallel beam search}.  The beam search refines these policies on the fly by pruning branches that are unfavourably evaluated by a \emph{discriminator}.  However, it can also harm \emph{diversity}, i.e., how well the agents cover the entire distribution of realistic behaviour, as pruning can encourage mode collapse.  Symphony addresses this issue with a hierarchical approach, factoring agent behaviour into \emph{goal generation} and \emph{goal conditioning}. The use of such goals ensures that agent diversity neither disappears during adversarial training nor is pruned away by the beam search. Experiments on both proprietary and open Waymo datasets confirm that Symphony agents learn more realistic and diverse behaviour than several baselines. 
\end{abstract}

%%%%%%%%%%%%%%%%%%%%%%%%%%%%%%%%%%%%%%%%%%%%%%%%%%%%%%%%%%%%%%%%%%%%%%%%%%%%%%%%
\section{INTRODUCTION}

Simulation is a crucial tool for accelerating the development of autonomous driving software because it can generate adversarial interactions for training autonomous driving policies, play out counterfactual scenarios of interest, and estimate safety-critical metrics.  In this way, simulation reduces reliance on real-world data, which can be expensive and/or dangerous to collect. As autonomous vehicles share public roads with human drivers, cyclists, and pedestrians, the underlying simulation tools require realistic models of these human road users. 

Such models can be obtained by applying \emph{learning from demonstration} (LfD) \cite{Argall2009-um,Hussein2017-yk} to example trajectories of human road use collected using sensors (e.g., cameras and LIDAR) already mounted on cars on the road.  Such demonstrations are ideally suited to learning the realistic road use behaviour needed for autonomous driving simulation.

However, existing LfD methods such as behavioural cloning \cite{Pomerleau1989-kb,michie1990cognitive} and generative adversarial imitation learning \cite{Ho2016-zi} are typically insufficient for producing realistic models of human road users. Despite minimising their supervised or adversarial losses, the resulting policies frequently collide with other road users or drive off the road.

To address this problem, we propose \emph{Symphony}, a new approach to LfD for autonomous driving simulation that can greatly improve the realism of the learned behaviour.
A key idea behind Symphony is to combine conventional policies, represented as neural networks, with a \emph{parallel beam search} that refines these policies on the fly.  As simulations are rolled out, Symphony prunes branches that are unfavourably evaluated by a \emph{discriminator} trained to distinguish agent behaviour from that in the data.  Because the beam search is parallelised, promising branches are repeatedly forked, focusing computation on the most realistic rollouts. In addition, since the tree search is also performed during training, the pruning mechanism drives the agent towards more realistic states that increasingly challenge the discriminator.  The results of each tree search can then be distilled back into the policy itself, yielding an adversarial algorithm. 

However, simply learning \emph{realistic} agents is not enough. They must also be \emph{diverse}, i.e., cover the entire distribution of realistic behaviour, in order to enable a full evaluation of autonomous driving software.  Unfortunately, while the use of beam search improves realism, it tends to harm diversity: repeated pruning can encourage mode collapse, where only the easiest to simulate modes are represented.  To address this issue, Symphony takes a hierarchical approach, factoring agent behaviour into \emph{goal generation} and \emph{goal conditioning}. For the former, we train a generative model that proposes goals in the form of routes, which capture high-level intent.  For the latter, we train goal-conditional policies that modulate their behaviour based on a goal provided as input.  Generating and conditioning on diverse goals ensures that agent diversity neither disappears during adversarial training nor is pruned away by the beam search.

We evaluate Symphony agents with extensive experiments on run segments from the Waymo Open Motion Dataset \cite{DBLP:journals/corr/abs-2104-10133} and a proprietary Waymo dataset consisting of demonstration trajectories and their corresponding contexts, created by applying Waymo's perception tools to sensor data collected by Waymo vehicles driving on public roads.  We report performance on several realism and diversity metrics, including a novel diversity metric called \emph{curvature Jensen-Shannon divergence} that indicates how well the high-level agent behaviour matches the empirical distribution. Our results confirm that combining beam search with hierarchy yields more realistic and diverse behaviour than several baselines.

\section{Background}

\subsection{Problem Setting}
\label{sec:ps}
The sequential process that generates the demonstration behaviour is multi-agent and general sum and can be modeled as a \emph{Markov game} \cite{littman1994markov} $\mathcal{G} = \{ \mathcal{S}, \mathcal{A}, P, \{r_i \} _{i=1}^N, \nu, \gamma \}$. $N$ is the number of agents; $\mathcal{S}$ is the state space; $\mathcal{A}$ is the action space of a given agent, (the same for all agents), such that the joint action and observation spaces are $\mathcal{A}^N$ and $\mathcal{Z}^N$; $\{r_i \} _{i=1}^N$ is the set of reward functions (one for each agent); $\nu$ is the initial state distribution; and $\gamma$ is a discount factor. The dynamics are described by a transition probability function $P(\boldsymbol{s}' | \boldsymbol{s}, \boldsymbol{a})$, where $\boldsymbol{a} \in \mathcal{A}^N$ is a joint action; $\boldsymbol{s} \in \mathcal{S}$ is also shown in bold because it factors similarly to $\boldsymbol{a}$.  The agent's actions are determined by a joint policy $\boldsymbol{\pi}(\boldsymbol{a}_t | \boldsymbol{s}_{t})$ that is agent-wise factored, i.e., $\boldsymbol{\pi}(\boldsymbol{a}_t | \boldsymbol{s}_{t}) = \prod_{i=1}^N\pi^i(a^i_{t} | \boldsymbol{s}_t)$.  

To avoid superfluous formalism, we do not explicitly consider partial observability.  However, this is easily modelled by masking certain state features in the encoders upon whose output the agents condition their actions.

Because we have access to a simulator, the transition function is considered known.  However, since we are learning from demonstration, the agents' reward functions are unknown.  Furthermore, we do not even have access to sample rewards.  Instead, we can merely observe the agents' behavior, yielding a dataset
$\mathcal{D}_E = \{\tau_k\}_{k=1}^{K_E}$  of $K_E$ trajectories where  $\tau_k = \{ (\boldsymbol{s}_{k,1}, \boldsymbol{a}_{k,1}), (\boldsymbol{s}_{k,2}, \boldsymbol{a}_{k,2}), \dots (\boldsymbol{s}_{k,T}, \boldsymbol{a}_{k,T}) \}$ is a trajectory generated by the `expert' joint policy $\boldsymbol{\pi}_E$.  The goal of LfD is to set parameters $\theta$ such that a policy $\boldsymbol{\pi}_\theta$ matches $\boldsymbol{\pi}_E$ in some sense. 

In our case, the data consists of LIDAR and camera readings recorded from an \emph{ego vehicle} on public roads.  It is partitioned into run segments and preprocessed by a perception system yielding the dataset $\mathcal{D}_E$.
The states $\bs{s}_{k,t}$ and discrete actions $\bs{a}_{k,t}$ in each run segment are fit greedily to approximate the logged trajectory.
Each state is a tuple containing three kinds of features.  The first is static scene features $s^{SS}_k$ such as locations of lanes and sidewalks, including a \emph{roadgraph}, a set of interconnected lane regions with ancestor, descendant, and neighbour relationships that describes how agents can move, change lanes, and turn. The second is dynamic scene features $s^{DS}_{k,t}$ such as traffic light states. The third is features describing the position, velocity, and orientation of the $N$ agents.  Together, these yield the tuple: $\boldsymbol{s}_{k,t} = \{ s^{SS}_k, s^{DS}_{k,t}, s^1_{k,t}, \ldots s^N_{k,t} \}$. By convention, $s^1_{k,t}$ is the state of the ego vehicle.  Since agents can enter or leave the ego agent's field of view during a run segment, we zero-pad states for missing agents to maintain fixed dimensionality.

To avoid the need to learn an explicit model of initial conditions, we couple each simulation to a \emph{reference trajectory}.  Each agent $i$ is initialised to the corresponding $s^i_{k,1}$ after which it can either be a \emph{playback agent} that blindly replays the behaviour in the reference trajectory or an \emph{interactive agent} that responds dynamically to the unfolding simulation using a policy $\pi^i_\theta$ learned from demonstration.  

During a Symphony simulation, the state of an interactive agent is determined by sampling actions from its policy $\pi^i_\theta$ and propagating it through $P$.  Given a reference trajectory $\tau$, this yields a new simulated trajectory $\tau'$ in which $s^{SS}_k$ and $s^{DS}_{k,t}$ remain as they were in the reference trajectory but the agent states $s^i_{k,t}$ of the $N_I$ interactive agents are altered.

\subsection{Behavioural Cloning}
\label{sec:bc}
The simplest approach to LfD is \emph{behavioural cloning} (BC) \cite{Pomerleau1989-kb,michie1990cognitive}, which solves a supervised learning problem: $\mathcal{D}_E$ is interpreted as a labeled training set and $\boldsymbol{\pi}_\theta$ is trained to predict $\boldsymbol{a}_t$ given $\boldsymbol{s}_t$.  If we take a maximum likelihood approach, then we can optimise $\theta$ as follows:
\begin{equation}
\label{eq:bc}
    \max_\theta \sum_{k=1}^{K_E}\sum_{t=1}^T\log\boldsymbol{\pi}_\theta(\boldsymbol{a}_{k,t}|\boldsymbol{s}_{k,t}).
\end{equation}
This approach is simple but limited.  As it optimises only the conditional policy probabilities, it does not ensure that the underlying distribution of states visited by $\boldsymbol{\pi}_E$ and  $\boldsymbol{\pi}_\theta$ match.  Consequently it suffers from \emph{covariate shift} \cite{Ross2011-bp}, in which generalisation errors compound, leading  $\boldsymbol{\pi}_\theta$ to states far from those visited by $\boldsymbol{\pi}_E$.

\subsection{Generative Adversarial Imitation Learning}
\label{sec:gail}
BC is a \emph{strictly offline} method because it does not require interaction with an environment or simulator.  Given $\mathcal{D}_E$, it simply estimates $\boldsymbol{\pi}_\theta$ offline using supervised learning.  By contrast, most LfD methods are \emph{interactive}, repeatedly executing $\boldsymbol{\pi}_\theta$ in the environment and using the resulting trajectories to estimate a gradient with respect to $\theta$.

Interactive methods include \emph{inverse reinforcement learning} \cite{Ng2000-wt,Abbeel2004-oi,ramachandran2007bayesian,Ziebart2008-ow} and adversarial methods such as \emph{generative adversarial imitation learning} (GAIL) \cite{Ho2016-zi}. GAIL borrows ideas from GANs \cite{Goodfellow2014-nm} and employs a \emph{discriminator} that is trained to distinguish between states and actions generated by the agents from those observed in $\mathcal{D}_E$. The discriminator is then used as a cost (i.e., negative reward) function by the agents, yielding increasingly log-like behaviour. In our multi-agent setting, the GAIL objective can be written as:
\begin{equation}
\label{eq:mgail}
  \min_{\boldsymbol{\theta}} \max_{\boldsymbol{\phi}} \Big[
  \bb{E}_{\boldsymbol{s}\sim d_\theta}\log(D_\phi(\boldsymbol{s}))  +
  \bb{E}_{\boldsymbol{s}^E\sim\mathcal{D}_E}\log(1-D_\phi(\boldsymbol{s}^E)) \Big],
\end{equation}
where $d_\theta$ is the distribution over states induced by $\boldsymbol{\pi}_\theta$ and $\mathcal{D}_E$ is here treated as an empirical distribution over states. Although the agents who generated $\mathcal{D}_E$ are not cooperative (as modelled by the different reward functions $\{r_i \} _{i=1}^N$ in $\mathcal{G}$), the learned agents controlled by $\boldsymbol{\pi}_\theta$ are cooperative because they all aim to minimise the same discriminator $D_\phi$, i.e., they share the goal of realistically imitating $\boldsymbol{\pi}_E$.

Differentiating through $d_\theta$ is typically not possible because $P$ is unknown.  Hence, updating $\theta$ requires using a score-function gradient estimator \cite{williams1992simple}, which suffers from high variance.  However, in our setting, $P$ is both known and differentiable, so we can employ \emph{model-based GAIL} (MGAIL) \cite{baram2017end}, which exploits end-to-end differentiability by directly propagating gradients from $D_\phi$ to $\boldsymbol{\pi}_{\theta}$ through $P$.

\section{SYMPHONY}

\begin{figure}
\centering
\includegraphics[width=1.03\columnwidth]{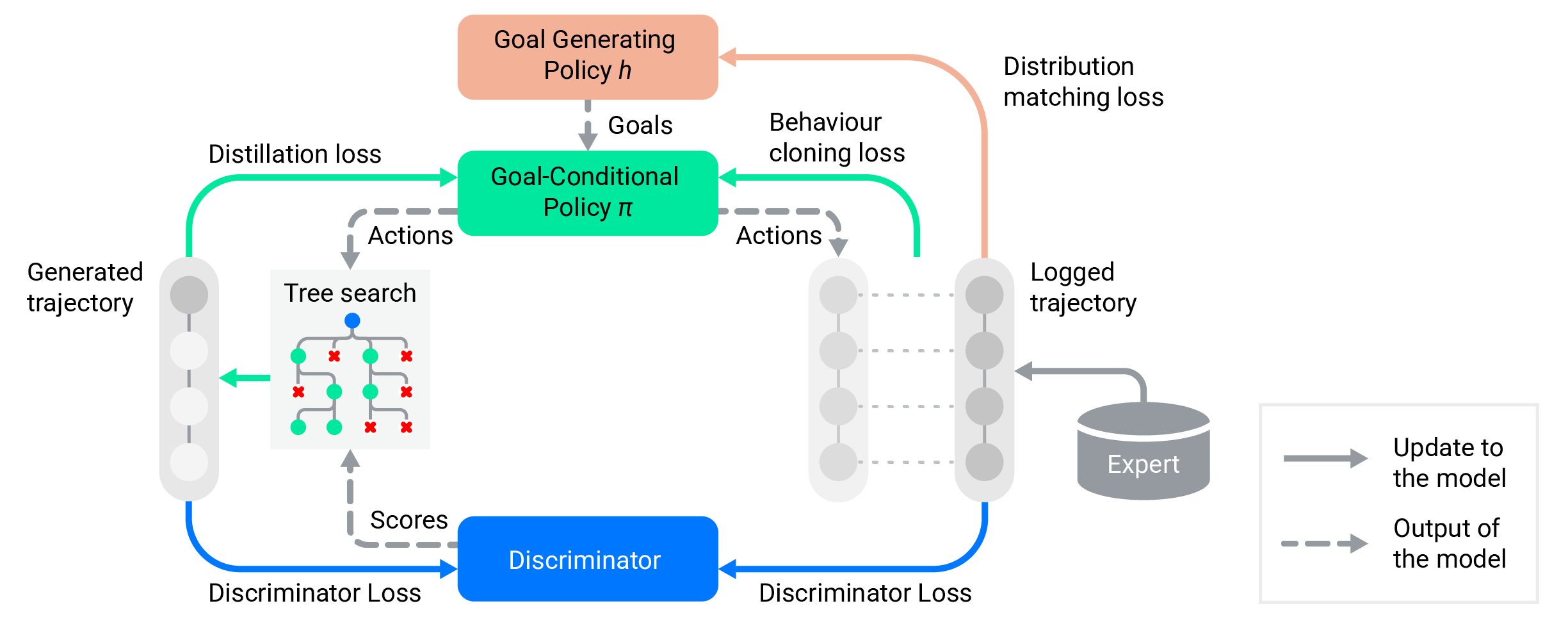}
\caption{Interactive agent training when Symphony is based on BC (replay agents not shown).}
%\vspace{-5mm} 
\label{fig:training}
\end{figure}

\begin{wrapfigure}{r}{0.55\columnwidth}
  \begin{center}
  \vspace{-6mm}
    \includegraphics[width=0.55\columnwidth]{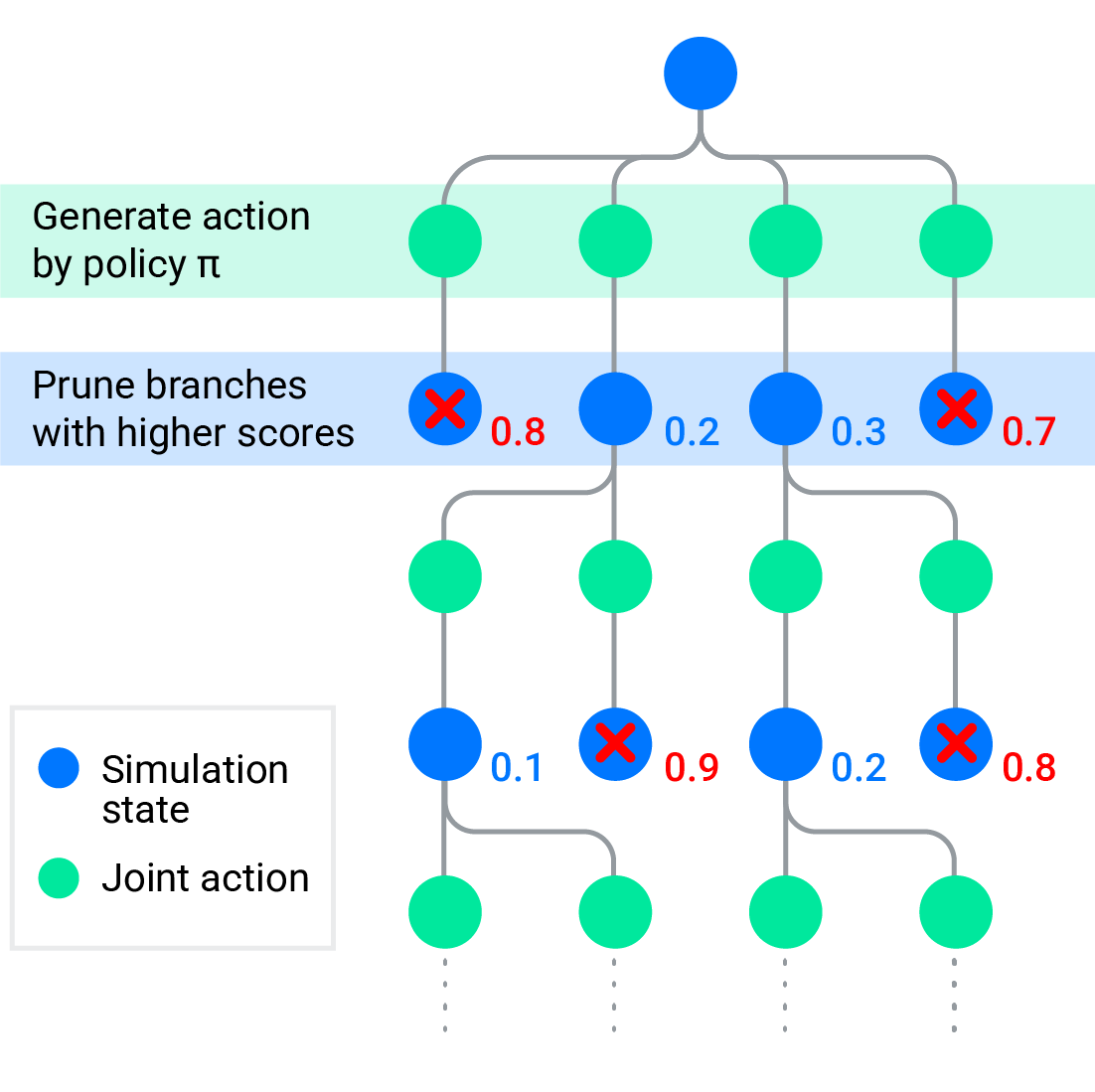}
  \end{center}
  \caption{Beam search.}
  \vspace{-3mm} 
 \label{fig:tree_search}
\end{wrapfigure}

Symphony is a new approach to LfD for autonomous driving simulation that builds upon a base method such as BC or MGAIL by adding a parallel beam search to improve realism and a hierarchical policy to ensure diversity.  Figure \ref{fig:training} gives an overview of the training process for the case where the base method is BC.  Training proceeds by sampling a batch of run segments from a training set $\mathcal{D}_{tr} \subset \mathcal{D}_E$ and using them as reference trajectories to perform new rollouts with the current policy.  At the start of each rollout, a goal generating policy proposes a goal, based on initial conditions, that remains fixed for the rollout and is input to the goal-conditional policy that proposes actions.  These actions are used to generate nodes in a parallel beam search (see Figure \ref{fig:tree_search}), which periodically prunes away branches deemed unfavourable by a discriminator and copies the rest to maintain a fixed-width beam. 

Unlike in model-predictive control \cite{garcia1989model} or reinforcement learning methods that employ online tree search, e.g.,  \cite{silver2016mastering,tesauro1996line}, in which an agent `imagines' various futures before selecting a single action, each rollout in Symphony is executed directly in the simulator.  However, because simulations happen in parallel, promising branches can be duplicated during execution to replace unpromising ones, focusing computation on the most realistic rollouts. 

Finally, we use the resulting rollouts to compute losses and update the goal generating policy, the goal-conditional policy, and the discriminator. During inference at test time, the process is the same except that run segments are sampled from a test set $\mathcal{D}_{te}$ and no parameters are updated.

In the rest of this section, we provide more details about the parallel beam search, hierarchical policy, network architectures, and learning rules.

\subsection{Parallel Beam Search}

For each reference trajectory in the batch, we first sample $S$ actions from the joint policy, yielding $S$ branches that roll out in parallel. We then call the discriminator at each simulation step to score each of the $N_I$ interactive agents in each branch, yielding a tensor of dimension $[B, N_I, T_p, S]$ where $B$ is the batch size and $T_p$ is the number of time steps between pruning/resampling. After every $T_p$ simulation steps, we aggregate the discriminator scores across the time and interactive agent dimensions, yielding a tensor of shape $[B, S]$ containing a score for each sample in the batch.  We aggregate by maximising across time and summing across agents. We then rank samples by aggregate score and prune away the top half (i.e., the least realistic). We tile the remaining samples such that $S$ remains constant throughout the simulation.  We use $T_p=10$, i.e., pruning and resampling occurs every 2 seconds of simulation.  During training, we use $S=4$ but during inference $S=16$.

Pruning based on aggregate scores means that the simulation at a given timestep can be subtly influenced by future events, i.e., actions are pruned away because they lead to unrealistic future states, and those states include observations of playback agents. In other problem settings, such as behaviour prediction \cite{lee2017desire,casas2018intentnet,DBLP:journals/corr/abs-1910-05449} such \emph{leakage} would be problematic because information about the future would not be available at inference time (the whole point is to predict the future given only the past).  However, in our setting the reference trajectory is available even at inference time, i.e., the goal is simply to generate realistic and diverse simulations given the reference trajectory.  While leakage can in principle yield useful hints, it can also be misleading as any leaked hints can become obsolete when interactive agents diverge from the reference trajectory.  In practice, as we show in Section \ref{sec:main-res}, refining simulation on the fly through beam search can drastically improve realism but tends to harm diversity: repeated pruning can encourage mode collapse, where only the easiest to simulate modes are represented.  Next we discuss a hierarchical approach to remedy this issue.

\subsection{Hierarchical Policy}

To mitigate mode collapse, we employ hierarchical agent policies. At the beginning of each rollout, a
high-level \emph{goal generating policy}  $h_{\bs{\psi}}(\bs{g}|\bs{s}_1)$ proposes a goal $\bs{g}$, based on an initial state $\bs{s}_1$, that remains fixed throughout the rollouts and is
provided to both the low-level \emph{goal-conditional} policy $\pi_{\bs{\theta}}(\bs{a}|\bs{s},\bs{g})$ and the discriminator
$D_{\bs{\phi}}(\bs{s} | \bs{g})$.
The goal generating policy is trained to match the distribution of goals in the training data:
\begin{equation}
\label{eq:goal_policy}
  \max_{\bs{\psi}} \mathbb{E}_{(\bs{s}^E_1, \bs{g}^E)\sim \mathcal{D}_E} \log h_{\bs{\psi}}(\bs{g}^E|\bs{s}^E_1).
\end{equation}
Because the same goal is used for all rollouts within the search tree, it cannot be biased by the discriminator.

We use routes, represented as sequences of roadgraph lane segments, as goals because they capture high-level intent and are a primary source of multi-modality.
A feasible set of routes is generated by following all roadgraph branches, beginning at the lane segment corresponding to the agent's initial state. 
From this set, routes with minimal displacement error from the observed trajectory are used as ground truth to train $h_{\bs{\psi}}(\bs{g}|\bs{s}_1)$ and as input to  $\pi_{\bs{\theta}}(\bs{a}|\bs{s}, \bs{g})$ during training. Hence, $h_{\bs{\psi}}(\bs{g}|\bs{s}_1)$ and $\pi_{\bs{\theta}}(\bs{a}|\bs{s}$ can be seen as learned versions of the router and planner, respectively, in a conventional control stack.

% \subsection{Interactive Agents}
\subsection{Architecture and Learning}

For each interactive agent, objects (such as other cars, pedestrians and cyclists), as well as static and dynamic features are all
encoded individually using MLPs, followed by max-pooling across inputs of the same type.
The resulting type-specific embeddings are, together with an encoding of features of the interactive agent, concatenated
and provided to the policy head as input.
Spatial information such as location or velocities of other objects are normalised with respect to the agent
before being passed into the network. Furthermore, roads and lanes are represented as a set of
points.
In large scenes, only the nearest 16 objects and 1$K$ static and dynamic features are included.  

For BC, the goal-conditional policy head maps the concatenated embeddings to a $7\times
21$ action space of discretised accelerations and steering angles. For MGAIL, we use a continuous action space specifying $x$-$y$ displacement to facilitate end-to-end differentiation. The goal generating policy maps to softmax logits for each
feasible route, up to a limit of 200 routes. The goal generating and goal-conditional policies use separate encoders. 
The discriminator uses a similar but simpler encoding by max-pooling across all objects and point features
within 20 metres.
We train both policies and the discriminator simultaneously.
We train the goal generating policy using \cref{eq:goal_policy} and the discriminator using:
\begin{equation*}
  \max_{\boldsymbol{\phi}} \Big[
  \bb{E}_{\boldsymbol{s}\sim d_{TS}}\log(D_\phi(\boldsymbol{s}))  +
  \bb{E}_{\boldsymbol{s}^E\sim\mathcal{D}_E}\log(1-D_\phi(\boldsymbol{s}^E)) \Big],
\end{equation*}
where $d_{TS}$ is generated by the tree search with $S$=4.
We train the goal-conditional policy using either BC or MGAIL.  In the case of BC, we increase its robustness to covariate shift by training not only on expert data, but also on additional data sampled from $d_{TS}$, i.e., the beam search is distilled back into the goal-conditional policy, yielding an adversarial method even without the use of MGAIL.
Each training batch contains 16 run segments of 10 seconds each, for which actions
are recomputed every 0.2 seconds.

\section{RELATED WORK}

\subsection{Coping with Covariate Shift}
One way to address covariate shift in BC is to add actuator noise when demonstrations are performed, forcing the demonstrator to label a wider range of states \cite{laskey2017dart}.  However, this requires intervening when demonstrations are collected, which is not possible in our setting.  Another solution is DAgger \cite{Ross2011-bp}, where the demonstrator labels the states visited by the agent as it learns, which also requires access to the demonstrator that is not available in our setting. Adversarial methods such as GAIL and MGAIL avoid covariate shift by repeatedly trying policies in the environment and minimising a divergence between the resulting trajectories and the demonstrations.  When the environment is not available, methods that match state distributions can retain BC's strictly offline feature while minimising covariate shift \cite{jarrett2020strictly}.  However, in our setting, remaining strictly offline is not necessary as we have access to a high quality simulator that is itself the target environment for the learned agents.

\subsection{Combining Planning and Learning}

When a model of the environment dynamics $P(s’ | s, a)$ is available, deliberative \emph{planning} can help to predict the value of different actions. Model-based reinforcement learning typically uses planning during training to reduce the variance of value estimates.  When the model is differentiable, this can also be exploited, as in MGAIL \cite{baram2017end,Baram2016-sq} to reduce the variance of gradient estimates \cite{Tamar2016-oq,lee2018gated,farquhar2018treeqn}.  By contrast, online planning typically uses tree search to refine policies on the fly during inference by focusing computation on the most relevant states \cite{tesauro1996line, sutton2018reinforcement}.  By distilling the results of the tree search back into the policy, online planning also serves as an extended policy improvement operation \cite{anthony2017thinking,silver2017mastering,silver2018general,brown2020combining,hamrick2020role,Schrittwieser_2020}. Recently, sequential decision making problems have been reformulated as auto-regressive models using transformer architectures \cite{chen2021decision}.  Most related to Symphony is the Trajectory Transformer \cite{janner2021reinforcement}, which is fully differentiable and uses beam search but without a discriminator.

\subsection{Autonomous Driving Applications}

As early as 1989, ALVINN \cite{Pomerleau1989-kb}, a neural network trained with BC, autonomously controlled a vehicle on public roads. More recently, deep learning has been used to train autonomous driving software end-to-end with BC \cite{Bojarski-nvidia} and perturbation-based augmentations have been used to mitigate covariate shift \cite{chauffeurnet-bansal}.
As simulation emerges as a crucial tool in autonomous driving, interest is turning to how to populate simulators with realistic agents.  ViBe \cite{Behbahani2018-wu} learns such models from CCTV data collected at intersections, using GAIL but without the tree search or hierarchical components of Symphony.  SimNet \cite{bergamini2021simnet} produces such models using only BC but uses GANs instead of reference trajectories to generate initial simulation conditions.  TrafficSim \cite{suo2021trafficsim} also uses hierarchical control like Symphony but with a latent variable model and without a tree search for online refinement. AdvSim \cite{DBLP:journals/corr/abs-2101-06549} is similar but generates adversarial perturbations to challenge the full autonomous driving stack.
Like Symphony, SMARTS \cite{zhou2020smarts} considers the realism and diversity of agents in a driving simulator, but employs only reinforcement learning, not LfD, to learn such agents. nuPlan \cite{DBLP:journals/corr/abs-2106-11810} is a planning benchmark that uses a set of reference trajectories but does not simulate agent observations, feeding the observations from the reference trajectory even if they diverge from the simulation.  

\section{EXPERIMENTS \& RESULTS}

\subsection{Experimental Setup}

\noindent {\bf Datasets.} We use two datasets.  The first, a {\em proprietary dataset} created by applying Waymo’s perception tools to sensor data collected by Waymo vehicles driving on public roads, contains $1.1M$ run segments each with 30$s$ of features at 15$Hz$. 
% Each segment has a corresponding scenario, such as changing lanes, merging, and includes object information interacting with a Waymo vehicle. 
The second dataset consists of $~64.5K$ run segments from the \emph{Waymo Open Motion Dataset} (WOMD) \cite{DBLP:journals/corr/abs-2104-10133}, which we extract to 10$s$ run segments sampled at 15$Hz$.\footnote{While we use the same run segments as the WOMD, states contain the features described in Section \ref{sec:ps}, not those in the WOMD.}  Both datasets exclude run segments containing more than 256 playback agents, 10$K$ roadgraph points, or fewer than $N_I$ agents at the initial timestep. Unlike the proprietary dataset, WOMD's run segments were selected to contain pairwise interactions such as merges, lane changes, and intersection turns.
Both datasets split the demonstration data $\mathcal{D}_E$ into disjoint sets $\mathcal{D}_{tr}$ and $\mathcal{D}_{te}$ with $K_{tr} = |\mathcal{D}_{tr}|$ and $K_{te} = |\mathcal{D}_{te}|$. For the proprietary dataset, $K_{tr}=1.1M$ and $K_{te}=10K$ and for the WOMD, $K_{tr}=58.1K$ and $K_{te}=6.4K$.

\vspace{0.1cm}\noindent
{\bf Simulation setup.} Unless stated otherwise, each simulation lasts for 10$s$, with initial conditions set by the reference trajectory and actions taken at 5$Hz$. Unless stated otherwise, the ego vehicle and one other vehicle are interactive, i.e., controlled by our learned policy, while the rest are playback agents.  The interactive agent is chosen heuristically depending on the context, e.g., in merges it is the vehicle with which the ego vehicle is merging.  If no such context applies, the nearest moving vehicle is chosen.

\subsection{Metrics}

We consider the following three realism metrics. {\bf Collision Rate} is the percent of run segments that contain at least one collision involving an interactive agent. A collision is detected when two bounding boxes overlap.
{\bf Off-road Time} is the percent of time that an interactive agent spends off the road.
{\bf ADE} is the average displacement error between each joint reference trajectory and the corresponding trajectory generated in simulation:
\begin{equation*}\label{eq:minSADE}
\textrm{ADE} = \frac{1}{K_{te}N_IT}\sum_{k=1}^{K_{te}} \Big[\sum_{i \in I} \sum^{T}_{t=1}  \delta(s^i_{k,t}, s^{'i}_{k,t}) \Big],
\end{equation*}
where $I$ is the set of indices of the interactive agents, $s^{'i}_{k,t}$ and $s^i_{k,t}$ are the states of the $i$th interactive agents in the $k$th simulated and reference trajectories respectively, and $\delta$ is a Euclidean distance function. 

We also consider two diversity metrics.
{\bf MinSADE} is the minimum scene-level average displacement error \cite{suo2021trafficsim}, which extends ADE to measure diversity instead of just realism. During evaluation, the simulator populates a set $\mathcal{R}_k$ by simulating $m$ trajectories for each reference trajectory $\tau_k$ in $\mathcal{D}_{te}$ and then computes minSADE as follows:
\begin{equation*}\label{eq:minSADE}
\textrm{minSADE} = \frac{1}{K_{te}N_IT}\sum_{k=1}^{K_{te}} \min_{\tau' \in \mathcal{R}_k} \Big[\sum_{i \in I} \sum^{T}_{t=1}  \delta(s^i_{k,t}, s^{'i}_{k,t}) \Big],
\end{equation*}
 
When $m=1$, minSADE reduces to ADE; when $m>1$, minimising minSADE requires populating $\mathcal{R}_k$ with diverse but realistic trajectories. Our experiments use $m=16$. Low minSADE therefore implies good coverage of behaviour modes.  However, it does not imply actually matching the empirical distribution of behaviours, e.g., low-probability modes may be over-represented. 
{\bf Curvature JSD} is a novel diversity metric that aims to measure how well the high-level behaviour matches the empirical distribution.  It is
computed using the roadgraph features in $s^{SS}_k$. Multiple lane regions that share a common ancestor are called \emph{branching regions} because they represent places where agents have multiple, branching choices, e.g., a lane approaching a four-way stop may branch into three descendant regions each going in a different direction at the intersection. For each branching region, we compute the average curvature across the region. The curvature JSD is then the Jensen-Shannon divergence between the distribution of average curvatures of the branching regions visited by the policy and reference trajectories. These distributions are approximated with histograms with bins of width $0.01$ in the range $[-1, 1]$, yielding $201$ bins.

\begin{table*}[]
\caption{Proprietary and Waymo Open Motion Dataset results and standard errors.}
\label{tab:main}
\renewcommand{\arraystretch}{1.2}
% \hspace{-8mm}
\resizebox{\textwidth}{!}{
\begin{tabular}{l|ccccc|ccccc}
\Xhline{2\arrayrulewidth}
         \multirow{3}{*}{Method} 
         & \multicolumn{5}{c|}{Proprietary Dataset} & \multicolumn{5}{c}{Waymo Open Motion Dataset} \\ \cline{2-11} 
         & \begin{tabular}[c]{@{}c@{}}Collision\\ rate (\%)\end{tabular} 
         & \begin{tabular}[c]{@{}c@{}}Off-road\\ time (\%)\end{tabular}  
         & \begin{tabular}[c]{@{}c@{}}ADE\\ (m)\end{tabular}  
         & \begin{tabular}[c]{@{}c@{}}MinSADE\\ (m)\end{tabular} 
         & \begin{tabular}[c]{@{}c@{}}Curvature JSD\\($\times10^{-3}$)\end{tabular} 
         & \begin{tabular}[c]{@{}c@{}}Collision\\ rate (\%)\end{tabular}  
         & \begin{tabular}[c]{@{}c@{}}Off-road\\ time (\%)\end{tabular}  
         & \begin{tabular}[c]{@{}c@{}}ADE\\ (m)\end{tabular}  
         & \begin{tabular}[c]{@{}c@{}}MinSADE\\ (m)\end{tabular}  
         & \begin{tabular}[c]{@{}c@{}}Curvature JSD\\($\times10^{-3}$)\end{tabular} \\
\Xhline{2\arrayrulewidth}
Playback & 0.99 & 0.40 & - & - & - 
         & 3.04 & 0.96 & - & - & - 
\\ \hline

BC & 16.65 {\scriptsize $\pm$0.41} & 2.16 {\scriptsize $\pm$0.08} & 5.80 {\scriptsize $\pm$0.07} & 2.16 {\scriptsize $\pm$0.04} & 2.82 {\scriptsize $\pm$0.26} & 24.65 {\scriptsize $\pm$0.32} & 2.75 {\scriptsize $\pm$0.12} & 4.81 {\scriptsize $\pm$0.10} & 1.76 {\scriptsize $\pm$0.04} & \bf 1.32 {\scriptsize $\pm$0.28} \\
BC + H & 17.25 {\scriptsize $\pm$1.07} & 1.69 {\scriptsize $\pm$0.12} & 5.18 {\scriptsize $\pm$0.09} & 2.01 {\scriptsize $\pm$0.03} & 1.38 {\scriptsize $\pm$0.11} & 23.27 {\scriptsize $\pm$0.38} & 2.40 {\scriptsize $\pm$0.05} & 4.38 {\scriptsize $\pm$0.04} & 1.67 {\scriptsize $\pm$0.03} & 3.50 {\scriptsize $\pm$1.25} \\
BC + TS & 1.84 {\scriptsize $\pm$0.13} & 0.35 {\scriptsize $\pm$0.03} & 4.83 {\scriptsize $\pm$0.09} & 2.07 {\scriptsize $\pm$0.04} & 5.28 {\scriptsize $\pm$1.14} & 4.94 {\scriptsize $\pm$0.65} & \bf 1.23 {\scriptsize $\pm$0.04} & 4.20 {\scriptsize $\pm$0.15} & 1.82 {\scriptsize $\pm$0.07} & 5.84 {\scriptsize $\pm$1.02} \\
BC + TS + H & \bf 1.80 {\scriptsize $\pm$0.07} & \bf 0.34 {\scriptsize $\pm$0.01} & \bf 4.30 {\scriptsize $\pm$0.05} & \bf 1.96 {\scriptsize $\pm$0.04} & \bf 1.24 {\scriptsize $\pm$0.14} & \bf 4.86 {\scriptsize $\pm$0.24} & 1.30 {\scriptsize $\pm$0.06} & \bf 3.70 {\scriptsize $\pm$0.09} & \bf 1.66 {\scriptsize $\pm$0.04} & 2.82 {\scriptsize $\pm$1.13} \\

% \Xhline{2\arrayrulewidth}
\hline
MGAIL & 5.34 {\scriptsize $\pm$0.32} & 0.83 {\scriptsize $\pm$0.08} & 7.32 {\scriptsize $\pm$0.52} & 3.95 {\scriptsize $\pm$0.42} & 1.88 {\scriptsize $\pm$0.18} & 9.48 {\scriptsize $\pm$0.91} & 1.62 {\scriptsize $\pm$0.15} & 5.70 {\scriptsize $\pm$0.44} & 3.13 {\scriptsize $\pm$0.26} & 4.14 {\scriptsize $\pm$1.36} \\
MGAIL + H & 4.16 {\scriptsize $\pm$0.18} & 0.76 {\scriptsize $\pm$0.03} & \bf 4.52 {\scriptsize $\pm$0.17} & \bf 2.48 {\scriptsize $\pm$0.09} & \bf 1.55 {\scriptsize $\pm$0.17} & 7.39 {\scriptsize $\pm$0.37} & 1.65 {\scriptsize $\pm$0.12} & \bf 3.82 {\scriptsize $\pm$0.12} & \bf 2.15 {\scriptsize $\pm$0.08} & 3.88 {\scriptsize $\pm$0.80} \\
MGAIL + TS & 2.97 {\scriptsize $\pm$0.16} & 0.72 {\scriptsize $\pm$0.20} & 6.83 {\scriptsize $\pm$0.48} & 3.80 {\scriptsize $\pm$0.39} & 4.14 {\scriptsize $\pm$1.08} & \bf 4.36 {\scriptsize $\pm$0.21} & \bf 1.22 {\scriptsize $\pm$0.02} & 4.26 {\scriptsize $\pm$0.19} & 2.51 {\scriptsize $\pm$0.11} & 5.42 {\scriptsize $\pm$1.26} \\
MGAIL + TS + H & \bf 2.40 {\scriptsize $\pm$0.19} & \bf 0.70 {\scriptsize $\pm$0.06} & 4.69 {\scriptsize $\pm$0.15} & 2.73 {\scriptsize $\pm$0.12} & 2.35 {\scriptsize $\pm$0.53} & 4.89 {\scriptsize $\pm$0.38} & 1.65 {\scriptsize $\pm$0.25} & 3.86 {\scriptsize $\pm$0.17} & 2.22 {\scriptsize $\pm$0.13} & \bf 2.80 {\scriptsize $\pm$0.60} \\

\Xhline{2\arrayrulewidth}
\end{tabular}}
% \vspace{-2mm}
\end{table*}

\begin{table*}[]
\caption{Proprietary dataset results and standard errors with 20 second rollouts and 8 interactive agents.}
 \label{tab:long-more}
\renewcommand{\arraystretch}{1.2}
% \hspace{-8mm}
\resizebox{\textwidth}{!}{
\begin{tabular}{l|ccccc|ccccc}
\Xhline{2\arrayrulewidth}
         \multirow{3}{*}{Method} 
         & \multicolumn{5}{c|}{Longer Rollouts (20 seconds)} & \multicolumn{5}{c}{More Interactive Agents ($N_I=8$)} \\ \cline{2-11} 
         & \begin{tabular}[c]{@{}c@{}}Collision\\ rate (\%)\end{tabular} 
         & \begin{tabular}[c]{@{}c@{}}Off-road\\ time (\%)\end{tabular}  
         & \begin{tabular}[c]{@{}c@{}}ADE\\ (m)\end{tabular}  
         & \begin{tabular}[c]{@{}c@{}}MinSADE\\ (m)\end{tabular} 
         & \begin{tabular}[c]{@{}c@{}}Curvature JSD\\($\times10^{-3}$)\end{tabular} 
         & \begin{tabular}[c]{@{}c@{}}Collision\\ rate (\%)\end{tabular}  
         & \begin{tabular}[c]{@{}c@{}}Off-road\\ time (\%)\end{tabular}  
         & \begin{tabular}[c]{@{}c@{}}ADE\\ (m)\end{tabular}  
         & \begin{tabular}[c]{@{}c@{}}MinSADE\\ (m)\end{tabular}  
         & \begin{tabular}[c]{@{}c@{}}Curvature JSD\\($\times10^{-3}$)\end{tabular} \\
\Xhline{2\arrayrulewidth}
Playback & 2.38 & 0.47 & - & - & - 
         & 2.37 & 0.22 & - & - & - 
\\ \hline

BC & 25.56 {\scriptsize $\pm$0.69} & 2.99 {\scriptsize $\pm$0.09} & 11.60 {\scriptsize $\pm$0.21} & 4.08 {\scriptsize $\pm$0.05} & \bf 2.61 {\scriptsize $\pm$0.20} & 17.46 {\scriptsize $\pm$0.79} & 1.06 {\scriptsize $\pm$0.15} & 6.40 {\scriptsize $\pm$0.51} & 1.86 {\scriptsize $\pm$0.14} & 1.49 {\scriptsize $\pm$0.25} \\
BC + H & 30.33 {\scriptsize $\pm$0.56} & 3.14 {\scriptsize $\pm$0.22} & 9.83 {\scriptsize $\pm$0.18} & 3.83 {\scriptsize $\pm$0.05} & 4.17 {\scriptsize $\pm$0.37} & 18.60 {\scriptsize $\pm$0.81} & 0.95 {\scriptsize $\pm$0.06} & 5.85 {\scriptsize $\pm$0.08} & \bf 1.83 {\scriptsize $\pm$0.08} & 1.73 {\scriptsize $\pm$0.36} \\
BC + TS & \bf 6.05 {\scriptsize $\pm$0.31} & 0.72 {\scriptsize $\pm$0.10} & 8.92 {\scriptsize $\pm$0.30} & 3.97 {\scriptsize $\pm$0.11} & 4.90 {\scriptsize $\pm$0.59} & \bf 4.17 {\scriptsize $\pm$0.23} & 0.38 {\scriptsize $\pm$0.02} & 6.22 {\scriptsize $\pm$0.38} & 2.10 {\scriptsize $\pm$0.20} & 2.76 {\scriptsize $\pm$0.50} \\
BC + TS + H & 7.76 {\scriptsize $\pm$0.12} & \bf 0.66 {\scriptsize $\pm$0.02} & \bf 7.74 {\scriptsize $\pm$0.15} & \bf 3.72 {\scriptsize $\pm$0.09} & 2.69 {\scriptsize $\pm$0.22} & 5.18 {\scriptsize $\pm$0.24} & \bf 0.36 {\scriptsize $\pm$0.03} & \bf 5.68 {\scriptsize $\pm$0.16} & 2.03 {\scriptsize $\pm$0.05} & \bf 0.96 {\scriptsize $\pm$0.12} \\

\hline

MGAIL & 14.15 {\scriptsize $\pm$0.83} & 1.03 {\scriptsize $\pm$0.19} & 14.18 {\scriptsize $\pm$0.97} & 9.34 {\scriptsize $\pm$0.45} & 6.79 {\scriptsize $\pm$1.70} & 11.08 {\scriptsize $\pm$1.29} & 0.47 {\scriptsize $\pm$0.02} & 12.30 {\scriptsize $\pm$0.62} & 6.55 {\scriptsize $\pm$0.65} & 4.14 {\scriptsize $\pm$1.06} \\
MGAIL + H & 14.73 {\scriptsize $\pm$1.93} & 1.72 {\scriptsize $\pm$0.24} & 10.52 {\scriptsize $\pm$1.17} & 6.58 {\scriptsize $\pm$0.60} & 3.09 {\scriptsize $\pm$0.62} & 6.79 {\scriptsize $\pm$0.11} & 0.37 {\scriptsize $\pm$0.03} & 5.74 {\scriptsize $\pm$0.34} & 2.80 {\scriptsize $\pm$0.26} & 1.61 {\scriptsize $\pm$0.40} \\
MGAIL + TS & 10.52 {\scriptsize $\pm$0.94} & 0.99 {\scriptsize $\pm$0.03} & 15.76 {\scriptsize $\pm$1.17} & 11.80 {\scriptsize $\pm$1.22} & 5.93 {\scriptsize $\pm$0.71} & 9.50 {\scriptsize $\pm$0.93} & 0.51 {\scriptsize $\pm$0.06} & 7.46 {\scriptsize $\pm$0.79} & 3.66 {\scriptsize $\pm$0.53} & 2.65 {\scriptsize $\pm$0.41} \\
MGAIL + TS + H & \bf 7.81 {\scriptsize $\pm$0.78} & \bf 0.88 {\scriptsize $\pm$0.03} & \bf 9.11 {\scriptsize $\pm$0.73} & \bf 6.39 {\scriptsize $\pm$0.64} & \bf 2.66 {\scriptsize $\pm$0.59} & \bf 5.80 {\scriptsize $\pm$0.47} & \bf 0.36 {\scriptsize $\pm$0.02} & \bf 5.47 {\scriptsize $\pm$0.07} & \bf 2.69 {\scriptsize $\pm$0.04} & \bf 1.24 {\scriptsize $\pm$0.15} \\

\Xhline{2\arrayrulewidth}
\end{tabular}}
%\vspace{-2mm}
\end{table*}

\subsection{Results}
\label{sec:main-res}

\begin{figure}[t]
\captionsetup[subfigure]{justification=centering}
  \begin{center}
    \subfloat[Proprietary Dataset.]{\label{fig:speed_dist}
      \includegraphics[width=0.47\columnwidth]{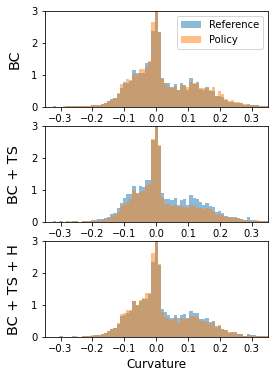} }
    \subfloat[Waymo Open Motion Dataset.] % Branching Region Curvature
    {\label{fig:diversity}
      \includegraphics[width=0.429\columnwidth]{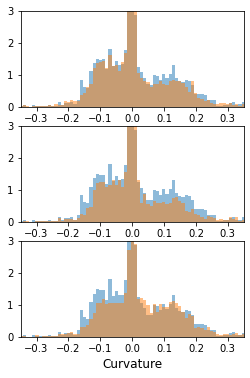} }
\end{center}
%\vspace{-0.2cm}
\caption{Histograms for distribution of curvature metrics.}
%\vspace{-0.4cm}
\label{fig:dist}
\end{figure}

We compare BC and MGAIL as is, with hierarchy (BC+H, MGAIL+H), with tree search (BC+TS, MGAIL+TS), and with both (BC+TS+H, MGAIL+TS+H).
We train each method for $70K$ update steps with run segments sampled uniformly from $K_{tr}$ and save checkpoints every $2K$ steps.  We then select the checkpoint with the lowest sum of collision rate and off-road time on a validation set of 200 run segments and test it using all of $K_{te}$. For each run segment in $K_{te}$, we generate 16 rollouts and report the average (or minimum for minSADE). We average all results over five independent seeds per method. For each metric, we indicate the best performing BC and MGAIL methods in bold. For collision rate and off-road time, we also report values for playing back the logs without interactive agents but computing these metrics for the agents that would have been interactive.  These values are slightly positive due to, e.g., perception errors on objects far from the ego vehicle or the use of bounding boxes instead of contours. 

Table \ref{tab:main} shows our main results, comparing all methods across all metrics on both datasets.  Comparing BC methods first, it is clear that tree search dramatically improves realism (especially with respect to collision rate and off-road time) but reduces diversity due to mode collapse.  This loss is detected by curvature JSD, which measures distribution matching, but not by minSADE, which only requires coverage.  However, the addition of hierarchy improves diversity in nearly all cases.  In particular, hierarchy is crucial for addressing the mode collapse from tree search.  BC+TS+H is the only BC method that gets the best of both worlds with strong performance on both realism and diversity metrics.

Figure \ref{fig:dist} shows the histograms used to compute the curvature JSD values in Table \ref{tab:main} for three BC methods, with learned policies in orange and the log reference trajectories in blue.  While BC matches distributions well, adding tree search leads to under-representation of positive curvature, i.e., right turns, a deficiency repaired with hierarchical policies.  %The effect is more pronounced on WOMD, which contains more intersection scenarios.

Turning now to MGAIL methods, similar trends emerge.  Tree search improves realism, though the effect is less dramatic as MGAIL is already adversarial even without tree search.  Similarly, while tree search also increases curvature JSD in MGAIL, the effect is smaller.   This is also to be expected given that MGAIL is already adversarial and can thus experience mode collapse even without tree search. On both datasets, the addition of hierarchy substantially improves MGAIL's diversity metrics.  Overall, the best BC methods perform better than the best MGAIL methods on nearly all metrics, though the differences are modest.

To see if we can maintain realism for longer time horizons, we repeat our experiments on the proprietary dataset with the rollout length doubled to 20$s$ in both training and testing.  The left side of Table \ref{tab:long-more} shows the results.  As expected, all methods obtain higher values on nearly all metrics in this more challenging setup.  However, the relative performance of the methods remains similar to that shown in Table \ref{tab:main}. Tree search methods perform much better with respect to realism than those without. While longer rollouts give more time to accumulate error, tree search repeatedly prunes problematic rollouts, greatly mitigating this effect.  BC+TS has worse curvature JSD than BC but the use of hierarchy prevents mode collapse, enabling BC+TS+H to approach the best of both worlds.  MGAIL shows less diversity loss from tree search than BC (only moderately worse minSADE) but also sees much better diversity when hierarchy is used. 

To assess whether we can maintain realism when more agents are replaced, we repeat our experiments on the proprietary dataset with eight interactive agents ($N_I=8$) in both training and testing. Two agents are selected as before and an additional six are selected that are nearest to the ego vehicle and whose distance traveled in the reference trajectory exceeds a threshold.  Again, all methods obtain higher values on most metrics, as with the longer rollouts discussed above.  Relative performance remains similar, with tree search improving realism but harming diversity and hierarchy improving diversity.  In this case, MGAIL sees no loss of diversity from tree search, as any mode collapse already happens in MGAIL training, but still sees substantial diversity improvements when hierarchy is used.

\section{CONCLUSIONS \& FUTURE WORK}

This paper presented Symphony, which learns realistic and diverse simulated agents and performs parallel multi-agent simulations with them. Symphony is data driven and combines hierarchical policies with a parallel beam search. Experiments on both open and proprietary Waymo data confirmed that Symphony learns more realistic and diverse behaviour than a number of baselines. 
Future work will investigate alternative pruning rules to shape simulation to various ends, augmenting goals to model driver persona, and developing additional diversity metrics that capture distributional realism in, e.g., agents' aggregate pass/yield behaviour.

\bibliography{icra-22-symphony}
\bibliographystyle{IEEEtran}

\end{document}